\documentclass[letterpaper, 10 pt, conference]{ieeeconf}  

\IEEEoverridecommandlockouts                              

\usepackage[T1]{fontenc}
\usepackage{url}
\usepackage{amssymb}
\usepackage{amsmath}
\usepackage{xcolor}
\usepackage{booktabs}
\usepackage{multirow}
\usepackage{graphicx}
\usepackage{capt-of}

\usepackage{pifont}
\newcommand{\cmark}{\textcolor{green!60!black}{\ding{51}}}
\newcommand{\xmark}{\textcolor{red!80!black}{\ding{55}}}

\title{\LARGE \bf
Tactile-JEPA: Topology-Aware Self-Supervised Representation Learning for Distributed Tactile Sensors
}

\author{Elizaveta~Kovtun$^{*,1,2}$, Matvey~Konovalov$^{1,3}$, Andrey~Sakhovskiy$^{1,2}$, and Semen~Budennyy$^{1,4}$
\thanks{$^{*}$Corresponding author: {\tt\small elizaveta.kovtun@skoltech.ru}}%
\thanks{$^{1}$Sber AI, Moscow, Russia}%
\thanks{$^{2}$Skolkovo Institute of Science and Technology (Skoltech), Moscow, Russia}%
\thanks{$^{3}$HSE University, Moscow, Russia}%
\thanks{$^{4}$Artificial Intelligence Research Institute (AIRI), Moscow, Russia}%
}

\begin{document}

\maketitle

\thispagestyle{empty}
\pagestyle{empty}


\begin{abstract}

Tactile sensing is an essential modality for robots performing contact-rich, dexterous manipulation, particularly under visual occlusion. While pre-trained image encoders are standard in robot learning pipelines, tactile encoders are still commonly trained from scratch from raw, noisy signals, which might limit their expressivity. Existing self-supervised learning (SSL) approaches focus predominantly on vision-based tactile sensors, leaving distributed electronic skins largely unaddressed. These sensors, however, have a distinctive property: their sensing elements are sparse and irregularly arranged over the surface they cover, which makes direct reuse of visual SSL methods suboptimal. We present Tactile-JEPA, an efficient self-supervised pre-training method that uses the spatial arrangement of tactile sensors to learn topology-aware representations. Specifically, it is trained to predict the embeddings of masked sensing elements from the unmasked remainder, using the sensor connectivity graph to guide spatial masking. Our analysis shows that effective tactile representations require capturing both local contact details and the global state of the tactile surface, which we achieve through dual-scale masking. Across three diverse datasets spanning magnetic and piezoresistive sensors, different robot embodiments, and single- and paired-sensor configurations, Tactile-JEPA reduces force estimation error by 6.3\% and in-hand orientation error by 20.8\% over the prior state-of-the-art, with consistent gains in other downstream applications, including policy learning. Overall, our results demonstrate that the benefit of tactile sensing depends critically on the quality of encoder pre-training, a problem which Tactile-JEPA addresses directly.
Code is available at \url{https://github.com/E-Kovtun/tactile}.
\end{abstract}

\section{Introduction}

People increasingly expect robots to assist them across diverse domains, from household tasks to manufacturing and agriculture~\cite{li2023behavior, keshvarparast2024collaborative, spagnuolo2025agricultural}, and this requires generalization across tasks and environments. Vision-Language-Action (VLA) models~\cite{kim2024openvla, black2024pi_0} provide a prominent foundation for this capability, but their only channel for world perception is vision. Relying on vision alone becomes a limiting factor under severe occlusions, dexterous contact-rich manipulation, and contact with fragile objects, in which \emph{tactile sensing} emerges as a critical modality~\cite{zhang2026tacvla, cheng2026omnivtla, li2026deco}. Recent works explore various ways of integrating touch into VLA learning pipelines~\cite{Li_2026_CVPR, bi2026vla}, demonstrating more precise manipulation in tactile-intensive scenarios. 


\begin{figure}[ht!]
  \centering
  \includegraphics[width=\columnwidth]{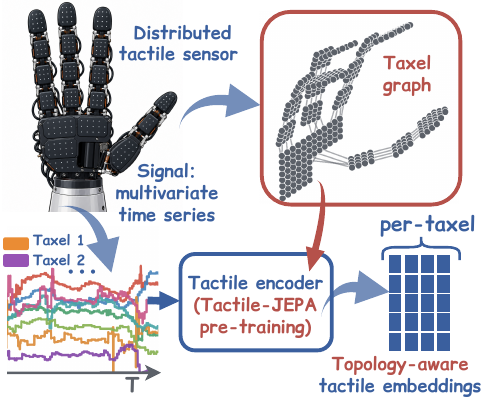}
  \caption{Tactile-JEPA pre-trains an encoder on signals from distributed tactile sensors in a self-supervised manner, accounting for sensor topology via the taxel graph and producing topology-aware per-taxel embeddings.}
  \label{fig:teaser}
\end{figure}

\begin{figure*}[ht!]
  \centering
  \includegraphics[width=\textwidth]{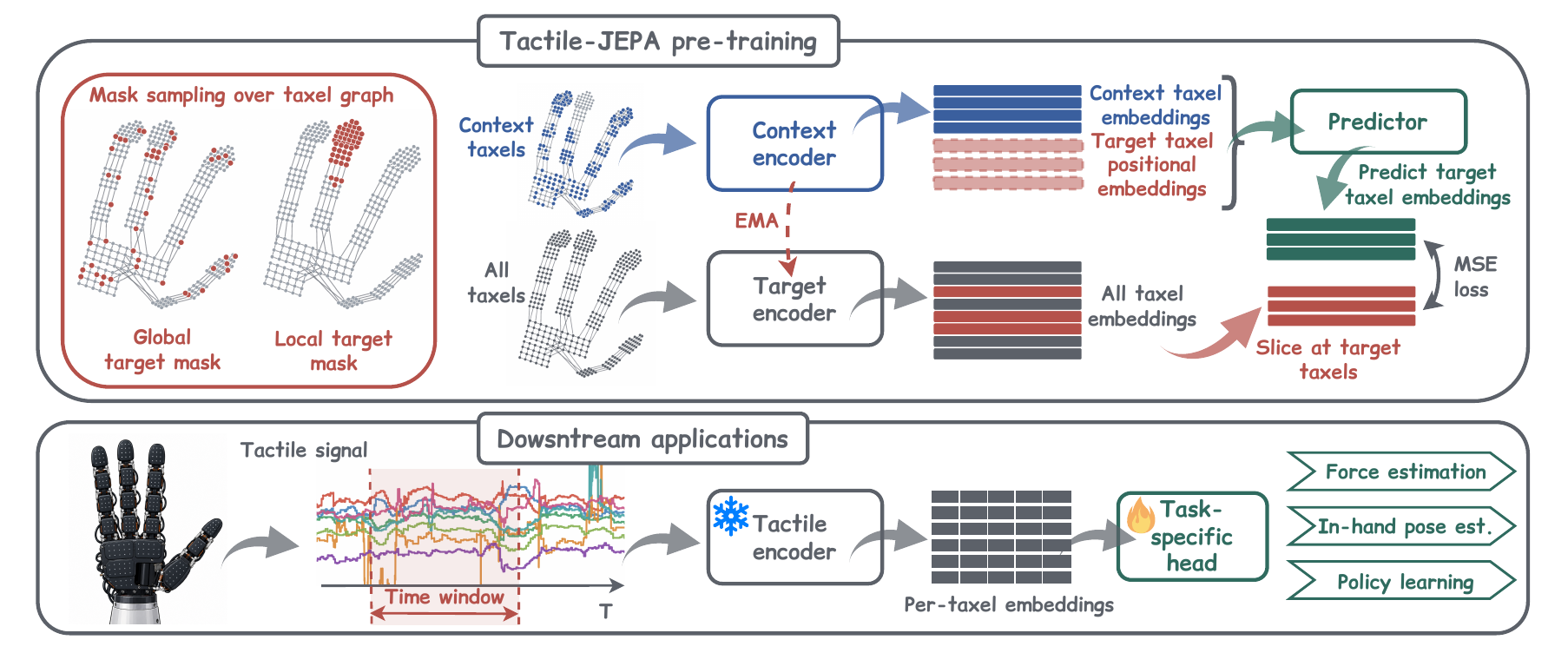}
  \caption{Overview of Tactile-JEPA. \textit{Top:} self-supervised pre-training. Target masks are sampled over the taxel graph at local and global scales (left), so the encoder learns topology-aware contact patterns at both extents. The predictor infers target embeddings from context embeddings, matching the EMA target encoder outputs via MSE loss (right). \textit{Bottom:} downstream applications. The target encoder from the top is frozen and reused as the tactile encoder, mapping a windowed tactile signal to per-taxel embeddings, on which a task-specific head is trained.}
  \label{fig:tactile-jepa}
\end{figure*}

Equipping robots with a sense of touch requires dedicated sensors, which differ substantially in their operating principle and design~\cite{lepora2026tactile}. Among the widely adopted are vision-based sensors, e.g., GelSight~\cite{yuan2017gelsight} and Digit~\cite{lambeta2020digit}, where an internal camera captures the deformation of a soft elastomer, producing an image stream that encodes contact geometry and forces. While this output is high-resolution and visually interpretable, such sensors are bulky and offer limited contact coverage~\cite{pmlr-v305-sharma25a}. An alternative is a thin and flexible electronic skin (e-skin)~\cite{cheng2019comprehensive, liu2022neuro}, which can be distributed across the entire hand or body rather than confined to fingertips. Regardless of the underlying transduction principle---piezoresistive \cite{luo2021learning}, capacitive (DexSkin~\cite{pmlr-v305-wistreich25a}), or magnetic (ReSkin~\cite{pmlr-v164-bhirangi22a}, Xela uSkin~\cite{tomo2016modular}, AnySkin~\cite{bhirangi2025anyskin})---e-skins output a multivariate time series whose channels are the readings of individual sensing elements, or taxels, distributed across the sensing surface~\cite{pmlr-v164-bhirangi22a, pmlr-v305-sharma25a}. A key question is how to effectively use the signal provided by the tactile sensors. While VLAs inherit vision and language representations from large-scale pre-trained backbones~\cite{beyer2024paligemma, wang2024qwen2}, the tactile modality has no such counterpart and is typically learned from scratch on raw, noisy data. This motivates pre-training tactile encoders to obtain effective representations~\cite{pmlr-v270-higuera25a, pmlr-v305-higuera25a}.

For vision-based tactile sensors, encoders are pre-trained by adapting self-supervised learning (SSL) approaches from computer vision, such as MAE~\cite{he2022masked} or DINO~\cite{caron2021emerging}. Beyond outperforming end-to-end training, the resulting representations transfer broadly: a single backbone serves different sensors and downstream tasks, remaining effective under limited labeled data~\cite{pmlr-v270-higuera25a, pmlr-v270-zhao25c}. The same strategy can be applied to e-skins, treating their multivariate time-series output as images~\cite{pmlr-v229-guzey23a}. However, such approaches do not account for the spatial structure of the sensing surface.
We close this gap with \textbf{Tactile-JEPA}, a self-supervised method for distributed tactile sensors. Following I-JEPA~\cite{assran2023self}, it predicts the embeddings of masked taxels from the visible ones. The novelty lies in mask sampling: masks are drawn (i) over the sensor connectivity graph, i.e., the taxel graph in Fig.~\ref{fig:teaser}, rather than a pixel grid, and (ii) at two scales, with \emph{local} masks covering compact regions and \emph{global} masks spanning taxels distributed across the skin. This produces topology-aware, multi-scale representations. Summing up, this work makes the following \textbf{contributions}:




\begin{itemize}
\item \textbf{Tactile-JEPA}, a self-supervised representation learning method for spatially distributed tactile sensors that operates directly on their multivariate time-series output and produces topology-aware, multi-scale representations.
\item Consistent gains over prior tactile representation learning methods across multiple embodiments, sensor types, and downstream tasks, including force and pose estimation, object and action classification, and tactile-conditioned policy learning.
\item An experimental study showing that masks sampled over the sensor connectivity graph outperform image-style blocks, and that their scale controls locality: local or global masks favor particular downstream tasks and, when mixed, transfer broadly.
\end{itemize}

\section{Related Work}

 
\subsection{Representation Learning for Vision-based Tactile Sensors}
Most representation learning techniques for tactile sensing are developed for vision-based sensors, whose raw output is an image, making vision SSL objectives directly applicable. A representative example is Sparsh~\cite{pmlr-v270-higuera25a}, a family of tactile encoders that transfer across several vision-based sensors and are pre-trained with a range of SSL objectives, including masked autoencoding (MAE~\cite{he2022masked}), self-distillation (DINO~\cite{caron2021emerging}, DINOv2~\cite{oquab:hal-04376640}), and joint-embedding prediction (I-JEPA~\cite{assran2023self}, V-JEPA~\cite{bardes2024revisiting}). Sparsh-X~\cite{pmlr-v305-higuera25a} extends this family to multisensory touch, using SSL to jointly encode tactile image, audio, inertial, and pressure channels into a single embedding. Beyond learning representations from the touch signal alone, another line of work builds a shared latent space over tactile, visual, and linguistic modalities, aligning them through contrastive pre-training that supports cross-modal tasks~\cite{yang2024binding, zhou2026collaborative, feng2025anytouch, ICLR2026_073c8584}. All of these methods inherit the pixel grid of vision-based sensors. Distributed sensors provide no such structure: their output is a multivariate time series over a sparse, irregular set of taxels.




\subsection{Representation Learning for Distributed Tactile Sensing}

A common workaround reshapes the distributed signal into an image, so that vision SSL remains applicable. T-DEX~\cite{pmlr-v229-guzey23a} pre-trains with BYOL~\cite{grill2020bootstrap} on magnetic Xela uSkin~\cite{tomo2016modular} pads distributed across a dexterous hand, arranging them into a single three-channel image. Sparsh-skin~\cite{pmlr-v305-sharma25a} abandons this image-like representation on the same hardware, encoding each taxel as a separate token trained via self-distillation~\cite{oquab:hal-04376640}. 
Signal structure is exploited spatio-temporally in STAT~\cite{lin2024jointly}, which combines masked reconstruction with time-order differentiation between signal segments.
In both Sparsh-skin and STAT, sensor topology enters only as a per-taxel location feature, leaving the connectivity of the sensing surface unrepresented. 
This issue is partially addressed in Tactile-GAT~\cite{chen2024tactile} and TacGNN~\cite{yang2023tacgnn}, which build adjacency graphs over taxels, yet both train end-to-end on labeled data, so no reusable representation is learned.  
HyperTaxel~\cite{li2024hypertaxel} does pre-train, but its contrastive objective requires contact surface geometry obtainable only in simulation. 
Consequently, existing approaches rely either on task-specific labels or privileged simulation information, neither of which is readily available in real-world deployment.
Therefore, the \textbf{research gap} we target is self-supervised pre-training that requires nothing beyond the sensor's own signal and its known layout, and that explicitly represents the sensor topology.

\begin{table*}[ht!]
\centering
\caption{Characteristics of open-source tactile datasets. Notation follows Sec.~\ref{problem}: $N$ taxels, $m$ sensing axes, $f_s$ sensor sampling rate, and $\tau=\lfloor T_w f_s \rfloor$ frames per time window. Coord.\ indicates whether taxel coordinates are provided. Hours and \#Frames give each dataset's total duration and frame count. $\tau$ matches human slip-response latency ($T_w \approx 0.1$\,s)~\cite{johansson1987signals}.}
\label{tab:datasets}
\setlength{\tabcolsep}{2.5pt}
\renewcommand{\arraystretch}{1.15}
\footnotesize
\begin{tabular*}{\textwidth}{@{\extracolsep{\fill}}llllllcrrrrrr@{}}
\toprule
\textbf{Dataset} & \textbf{Collection} & \textbf{Sensor} & \textbf{Type} &
\textbf{Platform} & \textbf{Setup} & \textbf{Coord.} &
$N$ & $m$ & $f_s$\,[Hz] & $\tau$ & \textbf{Hours} & \textbf{\#Frames} \\
\midrule
\textbf{Sparsh-skin~\cite{pmlr-v305-sharma25a}}
  & Teleop. play & Xela uSkin & Magnetic & Allegro hand & Unimanual & \cmark
  & 368 & 3 & 100 & 10 & 4.6 & 1.7\,M \\
\textbf{Tactile socks~\cite{luo2021learning}}
  & Human activity & Knitted fabric & Piezores. & Human feet & Bipedal & \cmark
  & 237 + 216 & 1 & 14 & 5 & 5.9 & 298\,k \\
\textbf{DECO-50~\cite{li2026deco}}
  & Teleop. demos & Inspire FTP & Piezores. & Inspire hands & Bimanual & \xmark
  & 1062 $\times$ 2 & 1 & 30 & 3 & 16.6 & 1.8\,M \\
\bottomrule
\end{tabular*}
\end{table*}

\section{Methodology}

\subsection{Problem Formulation} \label{problem}

Let a distributed tactile sensor covering a robot embodiment comprise $N$ taxels sampled at rate $f_s$. Over an observation window of duration $T_d$, it produces $T = \lfloor T_d f_s \rfloor$ frames forming the tactile signal $X = (x_1, \dots, x_N) \in \mathbb{R}^{N \times T \times m}$, where $x_i \in \mathbb{R}^{T \times m}$ the time series produced by the $i$-th taxel and $m$ is the number of sensing axes per taxel, e.g.\ $m = 1$ for pressure-based sensors, sensitive to the normal component only, or $m = 3$ for magnetic skins, whose $3$-axis magnetic flux readings respond to normal and shear components. 
A single frame captures the instantaneous state of the skin but carries no information about the contact dynamics. Therefore, we operate on short time windows of duration $T_w$,   comprising $\tau$ = $\lfloor T_w f_s \rfloor$ frames, and consider the slice $X_{\tau}^{(k)} = X[:, t_k:t_k+\tau, :] \in \mathbb{R}^{N \times \tau \times m}$. The value $T_w$ is chosen so that the window is not a noisy snapshot but a short history of the contact, yet still narrow enough to span a single event. We assume a known sensor layout, i.e., taxel arrangement. Taxel positions $P \in \mathbb{R}^{N \times T \times 3}$ may also be available, but Tactile-JEPA does not require them. 


Given an unlabeled dataset $\mathcal{D}=\{X_{\tau}^{(k)}\}_{k=1}^{K}$ of tactile windows, \textbf{our goal is to pre-train a tactile encoder $\mathbf{E}_{\theta}$ in a self-supervised manner that accounts for the geometric arrangement of the taxels}. The encoder maps each tactile window to a set of per-taxel embeddings. During downstream evaluation, shown at the bottom of Fig.~\ref{fig:tactile-jepa}, the pre-trained encoder $\mathbf{E}_{\theta}$ is kept frozen, and a lightweight task-specific head is trained on top of these embeddings using the corresponding labeled dataset. Importantly, the pre-training objective is defined within a short window and does not model temporal structure across windows, leaving temporal reasoning to the downstream decoder. Such a setup adds flexibility at inference: the encoder can be invoked at the control-loop rate, and longer history is obtained by composing successive embeddings.

\subsection{Tactile-JEPA: Self-Supervised Pre-Training} 
\label{tactile-jepa}




Tactile-JEPA is a self-supervised representation learning approach tailored to distributed tactile sensing. It learns to predict the representations of hidden taxel signals from visible ones in the embedding space, avoiding direct prediction of noisy sensor signals. 
The top of Fig.~\ref{fig:tactile-jepa} illustrates the Tactile-JEPA pre-training logic. It operates on two subsets of taxels: the visible ones, called the \emph{context}, and the hidden ones, called the \emph{targets}.
We call such a subset a \emph{region} and its binary indicator over all taxels a \emph{mask}, using the two terms interchangeably.
These regions are processed by three \textbf{components}.
The \emph{context encoder} $\mathbf{E}_{\theta}$ embeds only the context taxels. The \emph{target encoder} $\mathbf{E}_{\bar{\theta}}$ embeds all taxels, and its outputs at the target taxels serve as the prediction goals. The \emph{predictor} $\mathbf{P}_{\phi}$ receives the context embeddings together with the positions of the target taxels and predicts their embeddings. The self-supervised loss measures how closely these predictions match the target encoder outputs.

Below, we first describe how the tactile input signals are represented. We then introduce the \textbf{two key design elements} of Tactile-JEPA: the construction of the taxel connectivity graph and the mask sampling strategy based on this graph. Finally, we describe the Tactile-JEPA workflow, including the encoding, prediction, and SSL training objective.

\textbf{Input representation.} 
Let $x_i^{(k)} = X[i, t_k{:}t_k{+}\tau, :] \in \mathbb{R}^{\tau \times m}$ be the response of taxel $i$ over the $\tau$ frames. Following~\cite{pmlr-v305-sharma25a}, we map this response to an embedding of dimension $d$ by an affine projection shared across all taxels and all windows, followed by layer normalization (LN):
\begin{equation}
\tilde{x}_i^{(k)} = \operatorname{LN}\!\left(W \operatorname{vec}\big(x_i^{(k)}\big) + b\right) \in \mathbb{R}^{d},
\label{eq:tactile_input}
\end{equation}

where $\operatorname{vec}(\cdot)$ flattens the windowed signal of each taxel into a vector of length $m\tau$. The parameters $W \in \mathbb{R}^{d \times m\tau}$ and $b \in \mathbb{R}^{d}$ define the projection matrix and bias. We set the window duration $T_w$ to capture a single contact event. The full windowed response of a taxel therefore forms one semantic unit. The encoder maps this entire sequence to a single embedding. Stacking these embeddings across all taxels yields the representation $\tilde{X}_{\tau}^{(k)} \in \mathbb{R}^{N \times d}$. The dimension of this tensor depends strictly on taxel count and remains independent of the frame count within the window.

\textbf{Taxel connectivity graph.} 
We capture the spatial structure of the distributed tactile sensor with a taxel connectivity graph $G = (V, E)$, whose nodes $V = \{1, \dots, N\}$ correspond to the $N$ taxels and whose edges $(i,j) \in E$ connect taxels that are adjacent in the known sensor layout. The graph therefore requires no taxel coordinates and remains fixed across time windows. For sensors with two separate parts (e.g., two hands), $G$ consists of two connected components.

\textbf{Graph-based mask sampling.}
The SSL task divides the taxels into visible context regions and hidden target regions. The encoder predicts target representations from context representations. We sample $n_{c}$ context regions $\{\mathcal{C}_j\}_{j=1}^{n_{c}}$ and $n_{t}$ target regions $\{\mathcal{T}_l\}_{l=1}^{n_t}$, with $\mathcal{C}_j, \mathcal{T}_l \subset V$. Notably, Tactile-JEPA samples these regions on the taxel connectivity graph rather than on a pixel grid.
We define two types of target masks. A \emph{local} mask is a connected subgraph that covers a compact sensor region and captures localized contact patterns. Instead, a \emph{global} mask consists of taxels scattered across the graph, reflecting the overall contact state of the embodiment (e.g., the hand) rather than any single region. Both types are used jointly: each set of target masks contains an equal mix of local and global masks.
 Masks are assigned a budget, the number of taxels they contain. To sample a local mask, we choose a seed taxel at random and grow the mask by Dijkstra expansion,  repeatedly adding the connected taxels, until the budget is filled; a global mask is formed by selecting taxels uniformly at random over the whole graph until the same condition is met. The context mask follows the logic of the global mask construction. All target taxels are removed from the context to prevent overlap. For ratios and numbers of context and target masks, see Section~\ref{exp_setup}.



\textbf{Encoding.}
The tactile encoder $\mathbf{E}_{\theta}$ is a transformer operating over the prepared tactile representation $\tilde{X}_{\tau}^{(k)} \in \mathbb{R}^{N \times d}$. Before entering the transformer blocks, each taxel embedding $\tilde{x}_i^{(k)}$ is summed with a taxel positional embedding $e_i \in \mathbb{R}^d$ drawn from a learnable table $\{e_i\}_{i=1}^N$. 
Since each taxel contributes a single element to the sequence, bidirectional attention is computed across the taxel dimension, and the encoder outputs updated per-taxel embeddings $Z \in \mathbb{R}^{N \times d}$.

During pre-training, two instances of the same tactile encoder are used. The context encoder $\mathbf{E}_{\theta}$ processes only the taxel embeddings contained in a context region:

\begin{equation}
Z^{\mathcal{C}_j}_{\textit{context}} = \mathbf{E}_{\theta}\big(\{\tilde{x}_i^{(k)} + e_i\}_{i \in \mathcal{C}_j}\big) \in \mathbb{R}^{|\mathcal{C}_j| \times d}
\end{equation}

The target encoder $\mathbf{E}_{\bar{\theta}}$ has the same architecture but receives all taxels:

\begin{equation}
Z_{\textit{target}} = \mathbf{E}_{\bar{\theta}}\big(\{\tilde{x}_i^{(k)} + e_i\}_{i \in V}\big) \in \mathbb{R}^{N \times d}
\end{equation}

\textbf{Prediction.} 
The goal of the transformer-based predictor $\mathbf{P}_{\phi}$ is to infer the representations of a hidden target region from the representation of the observed context region. For a context--target pair $(\mathcal{C}_j, \mathcal{T}_l)$, the predictor receives the context representation $Z_{\textit{context}}^{\mathcal{C}_j}$ together with the positional embeddings of the taxels to be predicted, each summed with a learnable mask vector $u \in \mathbb{R}^d$:

\begin{equation}
\hat{Z}^{(\mathcal{C}_j, \mathcal{T}_l)} = \mathbf{P}_{\phi}\big(Z_{\textit{context}}^{\mathcal{C}_j}
\,\Vert\, \{e_i + u\}_{i \in \mathcal{T}_l}\big) \in \mathbb{R}^{|\mathcal{T}_l| \times d},
\end{equation}

where $\Vert$ denotes concatenation along the sequence dimension. Attention within $\mathbf{P}_{\phi}$ is computed over the concatenated sequence, and only the outputs at the target positions are retained, thereby determining the output dimension of $\hat{Z}^{(\mathcal{C}_j, \mathcal{T}_l)}$. 

\textbf{SSL objective.} For each context region, all target regions are predicted. Let $\hat{z}_i^{(j,l)}$ denote the prediction for taxel $i$ from $\hat{Z}^{(\mathcal{C}_j, \mathcal{T}_l)}$, and $z_i^{(l)}$ the representation of the same taxel $i$ obtained by slicing the target encoder output $Z_{\textit{target}}$ at the taxels of $\mathcal{T}_l$. The SSL objective of Tactile-JEPA is the mean squared error (MSE)  between them, averaged over all pairs:

\begin{equation}
\mathcal{L}_{\text{SSL}} = \frac{1}{n_c n_t}\sum_{j=1}^{n_c}\sum_{l=1}^{n_t} \frac{1}{|\mathcal{T}_l|} \sum_{i \in \mathcal{T}_l} \big\lVert \hat{z}_i^{(j,l)} - \operatorname{sg}(z_i^{(l)}) \big\rVert_2^2,
\label{eq:ssl_loss}
\end{equation}


Where $\operatorname{sg}(\cdot)$ denotes the stop-gradient operator. Backpropagation updates only the context encoder. The target encoder parameters are maintained as an exponential moving average (EMA) of the context encoder weights.

\section{Experiments and Results}

\subsection{Datasets}

We evaluate Tactile-JEPA on three publicly available tactile datasets, chosen to span distinct transduction principles, embodiments, and downstream tasks. We consider \textbf{Sparsh-skin}~\cite{pmlr-v305-sharma25a},
\textbf{Tactile socks}~\cite{luo2021learning}, and \textbf{DECO-50}~\cite{li2026deco}, which correspond to teleoperated play data collected with a sensorized robotic hand, human locomotion recorded from wearable tactile socks, and teleoperated demonstrations of contact-rich manipulation tasks on a bimanual robot, respectively. For DECO-50, we use only the data subset related to \emph{Assembly} task as the most tactile-intensive one~\cite{li2026deco}. The characteristics of the datasets are provided in Table~\ref{tab:datasets}.


\subsection{Downstream evaluation} \label{downstream}


\textbf{Tasks and metrics.} The Sparsh-skin dataset provides three downstream tasks~\cite{pmlr-v305-sharma25a}. In \emph{force estimation}, the model regresses tactile signals to 3-axis normal and shear forces. Labels are collected by indenting the palm sensor pad with a force/torque probe. We report the per-axis and total RMSE in centinewtons (cN). In \emph{in-hand pose estimation}, the model tracks the planar pose $(x, y, \theta) \in SE(2)$ of an object sliding under a static robotic hand. The pose is expressed in the hand frame and labeled using ArUco tags. We report the RMSE for $x$ and $y$ in centimeters (cm) and for $\theta$ in degrees. We also report pose accuracy, defined as the fraction of predictions within 2\,cm and 5$^{\circ}$ of the ground truth. \emph{Object classification} identifies the manipulated object from a set of 14 items in the play data. We evaluate this task using top-1 accuracy.

In the Tactile socks dataset, there are two downstream tasks~\cite{luo2021learning}. \emph{Action classification} predicts which of 9 activities the person wearing a pair of sensor socks is performing (e.g., walking, climbing up or down stairs, jumping) from a window of pressure frames; we evaluate it with top-1 accuracy. \emph{Full-body pose estimation} regresses the wearer's pose, represented as 19 relative joint angles spanning the legs, torso, and arms. We report mean joint RMSE in radians (rad) over the predicted joint angles.

On DECO-50, the goal is to \emph{learn a visuo-tactile manipulation policy} $\pi(A_t|O_t)$ that predicts a chunk of future actions $A_t = [a_t, a_{t+1}, \dots, a_{t+H}]$, where $H$ is the prediction horizon and each action $a_t \in \mathbb{R}^{12}$ specifies the target joint positions of the two dexterous hands ($6$ joints per hand), as commanded during bimanual teleoperation of the contact-rich Assembly task (plugging a socket held in one hand with a plug held in the other). The observation $O_t = [I_{1,t}, \dots, I_{n,t}, X_{\tau}^{(t)}]$ consists of the $n=2$ images from the binocular head camera at time $t$ together with the tactile history window $X_{\tau}^{(t)} = X[:, t-\tau:t, :] \in \mathbb{R}^{N \times \tau \times m}$ covering the $\tau$ frames preceding $t$. We report RMSE in normalized-action units between predicted and teleoperated actions, averaged over the $H=16$ horizon steps and 12 joints.

\textbf{Task heads.} We adopt the downstream head architectures of~\cite{pmlr-v305-sharma25a}, and train a separate head per task on top of the outputs of the frozen pre-trained tactile encoder.




\emph{Window-level head.} For force estimation and action and object classification, each target corresponds to one tactile window; for force, the window precedes the label, keeping the estimate causal. The head is an attentive probe, i.e., a one-layer, 3-head transformer whose learned query token cross-attends to the taxel embeddings and pools them into one embedding, followed by a two-layer MLP.

\emph{Sequence-to-sequence head.} Pose estimation on Sparsh-skin and Tactile socks requires longer temporal context, since pose is inferred from how contact evolves. We therefore encode consecutive windows with the frozen encoder, pool each with the attentive probe, and pass the resulting sequence to a one-layer transformer decoder that predicts the pose after each window.

\emph{Visuo-tactile policy.} For DECO-50, the embeddings pooled by the attentive probe, ResNet-18 visual embeddings, and a learnable action token form one sequence processed by a two-layer bidirectional transformer. A two-layer MLP maps the output action token to the action chunk.

\subsection{Baselines}
For a temporally matched comparison, we evaluate Tactile-JEPA against baselines pre-trained in the same regime as ours, on short time windows or instantaneous frames rather than on explicit long-horizon context.

\textbf{BYOL~\cite{pmlr-v229-guzey23a}.} T-DEX treats an instantaneous tactile frame as a three-channel image, with the sensing pads laid out spatially, and pre-trains an ImageNet-initialized AlexNet encoder on it with BYOL. In our setup, we feed it the middle frame of each time window. 

\textbf{MAE~\cite{pmlr-v305-sharma25a}.} Sparsh-skin (MAE) is trained by masked reconstruction of the taxel readings, with taxel coordinates concatenated to the input signal; otherwise, the tactile data view and the transformer encoder coincide with those of Tactile-JEPA.

\textbf{DINO~\cite{pmlr-v305-sharma25a}.} Sparsh-skin builds on the DINOv2 paradigm, where a student predicts the prototype logits of an EMA teacher from a more heavily masked view. Masking is taxel-wise; apart from concatenated taxel coordinates, the input data view and the encoder are the same as in Tactile-JEPA.

\textbf{End-to-end.} Without any pre-training, the tactile encoder is randomly initialized and trained jointly with the head on the labeled data of each task. On the tactile socks only, we replace the transformer with the CNN\&GRU encoder of~\cite{luo2021learning}.


\begin{table*}[ht!]
\caption{Comparison of the general-purpose quality of learned tactile representations across two distinct robotic embodiments and diverse downstream tasks. Best \textbf{bold}, second \underline{underlined}.}
\label{tab:metric_res}
\centering
\begingroup
\scriptsize
\setlength{\tabcolsep}{4pt}
\renewcommand{\arraystretch}{1.3}
\begin{tabular*}{\textwidth}{@{\extracolsep{\fill}}lllccccc@{}}
\toprule
\textbf{Dataset} &
\shortstack[l]{\textbf{Downstream}} &
\textbf{Metric} &
\textbf{End-to-end} &
\textbf{BYOL~\cite{pmlr-v229-guzey23a}} &
\textbf{MAE~\cite{pmlr-v305-sharma25a}} &
\textbf{DINO~\cite{pmlr-v305-sharma25a}} &
\shortstack{\textbf{Tactile-JEPA (ours)}} \\
\midrule

\multirow{11}{*}{\shortstack[l]{\textbf{Sparsh-skin~\cite{pmlr-v305-sharma25a}}}}
&  \shortstack[l]{Object classification}
& Accuracy $\uparrow$
& 70.61$\pm$2.85
& 64.30$\pm$1.38
& 63.44$\pm$0.64
& \underline{83.26$\pm$0.79}
& \textbf{83.27$\pm$3.36} \\

\cmidrule(lr){2-8}

& \multirow{4}{*}{\shortstack[l]{Force estimation}}
& RMSE [cN] $\downarrow$
& \underline{15.00$\pm$0.92}
& 23.79$\pm$0.96
& 21.55$\pm$2.06
& 16.66$\pm$1.17
& \textbf{14.05$\pm$0.73} \\
& & $x$-RMSE [cN] $\downarrow$
& \textbf{3.84$\pm$0.20}
& 5.70$\pm$0.38
& 5.68$\pm$0.57
& 4.23$\pm$0.18
& \underline{3.91$\pm$0.24} \\
& & $y$-RMSE [cN] $\downarrow$
& \underline{4.34$\pm$0.48}
& 5.55$\pm$0.32
& 5.69$\pm$0.78
& 4.46$\pm$0.67
& \textbf{4.23$\pm$0.40} \\
& & $z$-RMSE [cN] $\downarrow$
& \underline{25.32$\pm$1.57}
& 40.42$\pm$1.66
& 36.44$\pm$3.49
& 28.19$\pm$2.00
& \textbf{23.63$\pm$1.26} \\

\cmidrule(lr){2-8}

& \multirow{6}{*}{\shortstack[l]{In-hand pose est.}}
& $x$-RMSE [cm] $\downarrow$
& \underline{1.13$\pm$0.09}
& 1.31$\pm$0.13
& 1.60$\pm$0.30
& 1.20$\pm$0.08
& \textbf{0.96$\pm$0.05}$^{*}$ \\
& & $y$-RMSE [cm] $\downarrow$
& \underline{1.19$\pm$0.16}
& 1.36$\pm$0.05
& 1.42$\pm$0.19
& 1.31$\pm$0.02
& \textbf{1.06$\pm$0.04} \\
& & $\theta$-RMSE [${}^\circ$] $\downarrow$
& 7.20$\pm$0.87
& \underline{6.92$\pm$0.36}
& 8.38$\pm$0.70
& 8.27$\pm$0.44
& \textbf{5.48$\pm$0.39}$^{*}$ \\
& & $x$-Accuracy $\uparrow$
& \underline{91.93$\pm$2.05}
& 89.12$\pm$1.94
& 85.13$\pm$4.15
& 90.27$\pm$2.21
& \textbf{95.74$\pm$1.05}$^{*}$ \\
& & $y$-Accuracy $\uparrow$
& \underline{90.30$\pm$3.21}
& 86.97$\pm$0.87
& 87.12$\pm$2.54
& 88.95$\pm$1.03
& \textbf{93.14$\pm$0.99} \\
& & $\theta$-Accuracy $\uparrow$
& 62.85$\pm$6.24
& \underline{63.03$\pm$2.05}
& 53.54$\pm$3.10
& 54.02$\pm$1.84
& \textbf{71.75$\pm$2.56}$^{*}$ \\

\midrule

\multirow{2}{*}{\shortstack[l]{\textbf{Tactile socks~\cite{luo2021learning}}}}
& \shortstack[l]{Action classification}
& Accuracy $\uparrow$
& 92.97$\pm$1.26
& collapsed
& collapsed
& \underline{95.63$\pm$1.53}
& \textbf{96.91$\pm$1.92} \\

& \shortstack[l]{Full-body pose est.}
& \shortstack[l]{RMSE [rad] $\downarrow$}
& 14.37$\pm$0.30
& collapsed
& collapsed
& \underline{13.78$\pm$0.19}
& \textbf{13.42$\pm$0.33}$^{*}$ \\

\bottomrule
\end{tabular*}
\par\medskip
\begin{minipage}{\textwidth}
\scriptsize
$^{*}$Bold is statistically better than underlined ($p \leq 0.05$, one-sided Welch $t$-test).
\end{minipage}
\endgroup
\end{table*}

\begin{table}[t]
\caption{Comparison of tactile representations for visuo-tactile policy learning on DECO-50. Best \textbf{bold}, second \underline{underlined}.}
\label{tab:deco_res}
\centering
\setlength{\tabcolsep}{6pt}
\small
\begin{tabular}{lc}
\toprule
\textbf{Tactile representation} & \textbf{RMSE} $\downarrow$ \\
\midrule
Vision-only & 0.5132$\pm$0.0103 \\
Vision + Random embeddings & 0.4969$\pm$0.0118 \\
Vision + End-to-end tactile encoder & 0.4922$\pm$0.0058 \\
Vision + BYOL-pretrained encoder & collapsed  \\
Vision + MAE-pretrained encoder &  \textbf{0.4400$\pm$0.0021} \\
Vision + DINO-pretrained encoder & collapsed \\
\textbf{Vision + Tactile-JEPA encoder (ours)} & \underline{0.4634$\pm$0.0029} \\
\bottomrule
\end{tabular}
\par\smallskip
\begin{minipage}{0.95\linewidth}
\footnotesize
\end{minipage}
\end{table}

\subsection{Experimental Setup} \label{exp_setup}

\textbf{Data preprocessing.} For Sparsh-skin, magnetic readings are baseline-corrected by subtracting each taxel's mean response in a no-contact configuration, which removes the static offset~\cite{pmlr-v305-sharma25a}; the residuals are standardized per sensing axis. Piezoresistive skins need no such correction, as they rest at zero without contact. Tactile socks are therefore standardized with a single scalar mean and standard deviation, computed separately for each labeled subset, and DECO-50 signals are divided by a fixed per-hand maximum. All statistics are computed on training data only.

\textbf{Temporal windowing.} In Tactile-JEPA, we operate over short windows of tactile history rather than instantaneous frames. We set the number of frames per window $\tau$ so that $T_w \approx 0.1$\,s for Sparsh-skin and DECO-50 (Table~\ref{tab:datasets}), motivated by the latency at which humans adjust grip force following slip~\cite{johansson1987signals}. The tactile socks sample at $14$\,Hz, where this would reduce the window to a single frame; we therefore use $\tau=5$, giving $T_w \approx 0.36$\,s. For the pose estimation tasks, the sequence decoder consumes 10 consecutive windows on Sparsh-skin and 12 on the Tactile socks.

\textbf{Taxel grouping.} For the DECO-50 dataset, the two hands together contain $N = 1062 \times 2 = 2124$ taxels. Since our tactile encoder attends over taxels, such a sequence length is computationally expensive, so we group $n_g =4\text{--}5$ adjacent taxels and project their joint response $\mathbb{R}^{n_g \times \tau \times m}$ to a single embedding. This reduces the sequence length to $\approx N/n_g $ tokens.

\textbf{Mask sampling.} For Tactile-JEPA, we sample 1 context and 4 targets (2 local, 2 global), with 1 local and 1 global target per sensing part in two-part embodiments. The context initially covers 50–90\% of the taxels and each target 10–18\% of its sampling domain. Targets are drawn independently and may overlap; their union is removed from the context.

\textbf{Data splits.} For Sparsh-skin, object classification and pre-training share the same trajectories, split 80/10/10\% (train/val/test) within each object; test is held out. Force (64/16/20\%) and pose (71.7/18.3/10.0\%, trajectory-level) use separately collected data, disjoint from pre-training. For Tactile socks, action classification follows a chronological split (72.7/9.1/18.2\%) and pose follows the predefined splits (83.5/3.6/12.9\%). Encoders are pre-trained per task on the training portion. DECO-50 is split 70/15/15\% by complete demonstrations, with the training demonstrations shared between tactile encoder pre-training and policy learning.




\textbf{Architecture details.} In Tactile-JEPA, the encoder $\mathbf{E}_{\theta}$ follows the ViT-Tiny configuration~\cite{dosovitskiy2020image, touvron2021training}: 12 transformer blocks with embedding dimension $d=192$, 3 attention heads, and an MLP expansion ratio of 4. The predictor $\mathbf{P}_{\phi}$ uses the same configuration with only 4 blocks, keeping the information needed for prediction in the encoder representations rather than in the predictor.

\textbf{Training details.} Tactile-JEPA is trained with AdamW, a learning rate warmed up linearly from $10^{-5}$ to $10^{-4}$ and then cosine-decayed to $10^{-6}$, and weight decay increased from $0.04$ to $0.4$. Sparsh-skin and the Tactile-socks encoders are trained for 500 epochs at a global batch size of 256 (30 warm-up epochs); the DECO-50 encoder for 150 epochs at batch size 2048 (5 warm-up epochs). Downstream task heads are trained with AdamW at a learning rate of $10^{-4}$, with early stopping on validation and batch sizes of 4-64 depending on the task. The DECO-50 policy is trained for 150 epochs at batch size 1024. Full hyperparameters are available at \url{https://github.com/E-Kovtun/tactile}. We adapt the DINO and MAE configurations from Sparsh-skin,\footnote{\url{https://github.com/facebookresearch/sparsh-multisensory-touch}} the CNN\&GRU end-to-end baselines from the original Tactile socks implementation,\footnote{\url{https://github.com/YunzhuLi/senstextile}} and BYOL from T-DEX.\footnote{\url{https://github.com/irmakguzey/tactile-dexterity}}.

\textbf{Evaluation protocol.} For each SSL method, we pre-train three encoders with different seeds and train three downstream heads per encoder on Sparsh-skin and four on Tactile Socks and DECO-50, yielding 9 or 12 runs per task. End-to-end baseline follows three or four independent runs. We report means and sample standard deviations. The two top-ranked methods are compared with a one-sided Welch $t$-test on head-averaged results per pre-training seed ($n=3$; independent runs for non-pretrained baselines), at nominal $p \leq 0.05$ without multiple-comparison correction.

\textbf{Compute and runtime.} Tactile-JEPA pre-training uses four NVIDIA A100 80\,GB GPUs: 6.1\,h for Sparsh-skin, 0.7\,h and 2.0\,h for the Tactile socks action and pose encoders, and 5.5\,h for DECO-50. Downstream heads are trained on a single GPU, taking 0.6--2.1\,h, 0.2--0.4\,h, and 4.0\,h respectively.



\subsection{Results}

\textbf{Tactile representation evaluation.} We first evaluate whether Tactile-JEPA learns general-purpose tactile representations that transfer across downstream tasks. For each task, we train the task-specific heads described in Section~\ref{downstream} on the frozen representations from the Sparsh-skin and Tactile Socks datasets. Results are reported in Table~\ref{tab:metric_res}.
Compared with the strongest baseline on each metric, Tactile-JEPA reduces $\theta$-RMSE by 20.8\% and force RMSE by 6.3\%.
On Tactile socks, it improves action classification accuracy by 1.28\% and reduces full-body pose RMSE by 2.6\%.
Given the substantial and consistent margin over existing methods, we conclude that \emph{Tactile-JEPA learns representations that generalize across downstream tasks and embodiments}.

\textbf{Tactile-aware policy learning.}
We next investigate whether the learned tactile representations improve contact-rich policy learning on DECO-50 when integrated with visual information. We compare vision-only, vision augmented with random embeddings, a tactile encoder trained jointly with the policy from random initialization, and encoders pre-trained with SSL methods. Results are given in Table~\ref{tab:deco_res}. 
MAE achieves the lowest error on DECO-50 but does not maintain consistent superiority across datasets. Tactile-JEPA achieves the second-best results with stable training and no adaptation, while DINO collapses in its standard configuration and becomes viable only after disabling iBOT loss.



Across three datasets spanning different distributed tactile sensors, Tactile-JEPA shows \emph{robust training dynamics on heterogeneous tactile signals}: it pre-trains stably on all of them under the same masking configuration, with no per-dataset tuning. In contrast, several baselines exhibit training instability: BYOL collapses on both Tactile Socks and DECO-50, while MAE collapses on Tactile Socks.


\textbf{Computational efficiency.} Tactile-JEPA pre-trains 65--75\% faster than DINO across all datasets while matching or exceeding its downstream performance.


\section{Ablation Study}

\textbf{Masking strategy ablation.} Since Tactile-JEPA derives its topology awareness and multi-scale representations from sampling target masks over the connectivity graph, we primarily ablate the target-mask sampling strategy. In particular, we compare block-based masking from I-JEPA~\cite{assran2023self} (4I-JEPA), four local masks (4L), four global masks (4G), and our default mix of two local and two global (2L+2G). Results are provided in Table~\ref{tab:mask_ablation}.
As local targets focus on fine contact detail and global ones on the whole-hand contact state, single-scale targets can favor specific tasks. On DECO-50, 4G yields the lowest policy error, consistent with action prediction depending on the contact state of the whole hand rather than on any single contact region. In contrast, our default 2L+2G mix achieves the best force RMSE and in-hand $x$-accuracy, ranks second on DECO-50 close to 4G, and outperforms graph-agnostic I-JEPA masking on all three tasks. \emph{Multi-scale targets thus provide a robust choice for general-purpose pre-training}.

We additionally analyze the default globally sampled context mask with a locally sampled one ($^{\dagger}$ in Table~\ref{tab:mask_ablation}). The multi-scale target setup 2L+2G with local context improves force estimation but degrades pose and policy performance substantially, confirming global context as the better general-purpose choice. Notably, 4L target masks with local context outperform I-JEPA masking on both force estimation and policy learning. Since both use compact, connected masks with the same budget and differ only in whether masks are sampled over the taxel graph, this comparison isolates and confirms the benefit of graph-based sampling.

\begin{table}[t]
\caption{Evaluation of different target-mask sampling strategies for Tactile-JEPA on Sparsh-skin (force and in-hand pose est.) and DECO-50. Best \textbf{bold}, second \underline{underlined}.}
\label{tab:mask_ablation}
\centering
\setlength{\tabcolsep}{4pt}
\small
\begin{tabular}{lccc}
\toprule
& \textbf{Force est.} & \textbf{In-hand pose est.} & \textbf{Policy learning} \\
\cmidrule(lr){2-2}\cmidrule(lr){3-3}\cmidrule(lr){4-4}
\textbf{Targets} &
\textbf{RMSE $\downarrow$} &
\textbf{$x$-Acc. (\%) $\uparrow$} &
\textbf{RMSE $\downarrow$} \\
\midrule

4I-JEPA & \underline{14.99}$\pm$1.03 & 93.50$\pm$1.46 & 0.4959$\pm$0.0083 \\

4L & 15.52$\pm$1.03 & 94.32$\pm$1.86 & 0.4705$\pm$0.0145 \\
4G & 15.17$\pm$0.96 & \underline{95.32$\pm$1.65} & \textbf{0.4600$\pm$0.0019} \\

\textbf{2L+2G} & \textbf{14.05$\pm$0.73} & \textbf{95.74$\pm$1.05} & \underline{0.4635$\pm$0.0044} \\

\midrule

4L$^{\dagger}$& 13.93$\pm$1.03 & 91.29$\pm$1.49 & 0.4893$\pm$0.0130  \\
4G$^{\dagger}$& 15.14$\pm$1.00 & 95.16$\pm$0.73 & 0.4807$\pm$0.0078  \\
2L+2G$^{\dagger}$& 13.28$\pm$0.83 & 92.83$\pm$2.17 & 0.4828$\pm$0.0102  \\

\bottomrule
\end{tabular}
\par\smallskip
\begin{minipage}{0.95\linewidth}
\footnotesize
$^{\dagger}$Context mask sampled locally instead of globally (default); excluded from the best and second-best marking.
\end{minipage}
\end{table}

\section{Conclusion}

We introduce Tactile-JEPA, a self-supervised pre-training method that exploits the spatial arrangement of taxels in distributed tactile sensors. The key trait of Tactile-JEPA is multi-scale mask sampling over the taxel graph, which makes the representations topology-aware, encoding both localized tactile cues and the overall contact state of the hand. Since the taxel graph is used only for mask sampling during pre-training, the encoder requires no layout information in downstream tasks. The resulting embeddings are general-purpose, transferring across task types and embodiments, and reduce error over the strongest baselines by 20.8\% in in-hand orientation estimation and 6.3\% in force estimation. Moreover, Tactile-JEPA pre-trains stably across all sensor types, whereas competing SSL methods collapse on some datasets. Overall, Tactile-JEPA brings effective wide-coverage tactile sensing within practical reach of robot learning.

\section*{ACKNOWLEDGMENT}

The authors used an LLM for language editing and grammar checking, and take full responsibility for all content.

\bibliographystyle{IEEEtran}
\bibliography{IEEEabrv, IEEEexample}

\end{document}